\documentclass{article}
\usepackage[accepted]{icml2025}

\usepackage{microtype}
\usepackage{graphicx}
\usepackage{booktabs}
\usepackage{amsmath,amssymb,amsthm}
\usepackage{mathtools}
\usepackage{filecontents}
\usepackage{tikz}
\usetikzlibrary{positioning,trees,matrix}

\theoremstyle{plain}

\theoremstyle{definition}

\theoremstyle{remark}

\icmltitlerunning{Working Title}

\begin{document}

\twocolumn[
\icmltitle{Natural Language Edge Labelling: Decoupling Intent from Execution in Structured LM Reasoning}

\begin{abstract}
Controllers for structured LM reasoning---e.g., Chain-of-Thought, self-consistency, and Tree-of-Thoughts---often entangle \emph{what to try next} with \emph{how to execute it}, exposing only coarse global knobs and yielding brittle, compute-inefficient, and hard-to-audit behavior. We introduce \textbf{Natural Language Edge Labelling (NLEL)}, a labeller--tuner overlay that attaches a free-form natural-language directive to each search edge and translates it into a \emph{schema-bounded control vector} for decoding, search (branch quotas, exploration $\beta$), generation bundle size, retrieval mixtures, and verification passes. A labeller $\Lambda$ emits labels from the parent state and a compact context; a tuner $\Psi$ maps $(P, L, C)\to \Pi$, with strict schema validation and \emph{trust-region projection} around safe defaults. Downstream selection remains ToT-style with score $S=\mu+\beta\sigma$ and depth-annealed $\beta$. We show NLEL strictly generalizes CoT/ToT, prove an anytime-monotonicity property for top-$k$ selection under label-conditioned bundles, and bound selector shortfall by control-vector distortion---providing decision-relevant justification for guards like trust regions and verification passes. We instantiate $\Psi$ as a prompt-only \emph{JSON Parameter Emitter} and preregister an evaluation on GSM8K, MATH (subset), StrategyQA, and ARC-Challenge with compute-aware reporting (success@compute, tokens-per-success) and ablations over $\Lambda$, $\Psi$, trust-region radius, and control quantization; preregistered forecasts anticipate accuracy gains at comparable token budgets and improved success@compute under constraints. NLEL offers an interpretable, model-agnostic interface that separates intent from execution for controllable, auditable LM inference.
\end{abstract}

\begin{icmlauthorlist}
\icmlauthor{Abhinav Madahar}{ind}
\end{icmlauthorlist}
\icmlaffiliation{ind}{Independent Computer Scientist}

\icmlcorrespondingauthor{Abhinav Madahar}{abhinavmadahar@gmail.com}
]


\section{Introduction}

Structured reasoning with large language models (LMs) has advanced rapidly through prompting and search-based controllers such as Chain-of-Thought (CoT), self-consistency, and Tree-of-Thoughts (ToT). These methods elicit multi-step ``thinking'' and expand multiple partial solutions, but they typically couple \emph{what to try next} with \emph{how to run it} and expose only coarse global knobs (e.g., temperature, beam size). As a result, inference-time behavior can be brittle, compute-inefficient, and hard to audit: the controller rarely explains the intended operation for each step, and low-level decoding, search, retrieval, and verification settings are adjusted only indirectly.

We propose \textbf{Natural Language Edge Labelling (NLEL)}, a control layer for structured LM reasoning that decouples \emph{semantic intent} from \emph{execution control}. Each edge in the search structure carries a free-form natural-language label $L$ (e.g., ``seek a counterexample,'' ``work backward from the goal,'' ``call retrieval; summarize before deciding''), produced by a \emph{labeller} LM \(\Lambda\) from the current parent state \(P\) and a compact context \(C\). A \emph{tuner} LM \(\Psi\) then maps the triple \((P,L,C)\) into a schema-bounded \emph{control vector} \(\Pi\) that configures decoding (temperature, top-\(p\), max tokens), search (branch quota, exploration coefficient \(\beta\)), generation bundle size, retrieval mixture weights \(w\), and verification passes. Downstream selection is ToT-style using the score \(S=\mu+\beta\sigma\) with depth-annealed \(\beta\). To stabilize behavior, the emitted \(\Pi\) is validated against a schema and projected into a trust region around safe defaults \(\Pi_{0}\). This overlay is \emph{model-agnostic}: the child reasoner and the ToT selector remain unchanged.

Our analysis shows that NLEL strictly generalizes CoT/ToT controllers (recovering them as special cases) and that enlarging the candidate pool with label-conditioned bundles yields \emph{anytime monotonicity} under the ToT selector (top-\(k\) \(S\) cannot decrease when adding labelled candidates). We further derive bounds linking control distortion to selector shortfall (a ``bits-to-performance'' statement) and justify guards such as trust-region projection and verification passes. Figure-style synthetic traces illustrate label semantics paired with tuned controls across a simple proof.

Empirically, we evaluate NLEL as a \emph{prompt-only JSON Parameter Emitter (JPE)} instantiation of \(\Psi\), paired with \(\Lambda\), on four public suites---GSM8K, MATH (subset), StrategyQA, and ARC-Challenge---using compute-aware metrics (success@compute; tokens-per-success). We preregister forecasts indicating accuracy gains over strong ToT-style baselines at comparable token budgets and improved success@compute in constrained regimes; full measured results will replace forecasts once runs complete.

\paragraph{Contributions.}
\begin{enumerate}
\item \textbf{Method:} A labeller--tuner overlay that translates free-form edge labels into schema-bounded control vectors for decoding, search, retrieval, and verification, with trust-region and budget guards.
\item \textbf{Theory:} Reductions showing that NLEL recovers CoT/ToT; an anytime monotonicity result under \(S=\mu+\beta\sigma\); bounds connecting control distortion to selector shortfall; and simple compute accounting under quotas.
\item \textbf{Instantiation:} A prompt-only JPE realizing \(\Psi\) with a compact in-prompt ledger and normalized schema.
\item \textbf{Evaluation plan:} Pre-registered experimental design across four benchmarks with compute-aware reporting and ablations isolating \(\Lambda\), \(\Psi\), trust-region radius, and control quantization.
\end{enumerate}

\section{Related Work}

Structured reasoning with language models has become a prominent research direction. One foundational approach is \emph{Chain-of-Thought (CoT) prompting}, which elicits intermediate reasoning steps from the model. By having the model ``think aloud'' through a series of sub-steps, CoT significantly improves performance on complex tasks~\cite{wei2022chain}. Building on this idea, more advanced frameworks allow branching and search over possible reasoning paths instead of following a single linear chain. For example, \emph{Tree-of-Thoughts (ToT)} generalizes CoT by expanding a tree of potential ``thought'' steps and using self-evaluation to decide among branches, enabling lookahead and backtracking during inference~\cite{yao2023tree}. Such deliberate search over a reasoning tree or DAG can yield far better results on tasks requiring planning or strategic exploration, as the model is not confined to greedy left-to-right generation. Relatedly, alternative decoding strategies like self-consistency have been proposed to improve reliability: instead of taking one pass through the prompt, the model samples multiple diverse reasoning chains and then selects the most consistent answer among them~\cite{wang2022selfconsistency}. This method leverages the intuition that a complex problem may be solved via different logical routes leading to the same answer, and indeed has been shown to greatly boost accuracy on benchmarks~\cite{wang2022selfconsistency}.

Another line of work augments language model reasoning with \emph{external tools or formal executors}. \emph{Program-Aided Language Models (PAL)} exemplify this trend by generating programs (e.g., Python code) as intermediate reasoning steps and offloading their execution to a runtime~\cite{gao2023pal}. In PAL, the language model’s job is to correctly decompose the problem into code, and the Python interpreter handles the actual calculation or logic—an approach that achieved state-of-the-art results on math and symbolic reasoning tasks, even outperforming much larger models that rely on CoT alone~\cite{gao2023pal}. Beyond code execution, researchers have equipped LMs with a variety of \emph{tool-use abilities}. \emph{Toolformer} showed that an LM can be fine-tuned (in a self-supervised way) to decide when to call external APIs (such as calculators, web search, or translation services) and how to incorporate the results into its text generation~\cite{schick2023toolformer}. This allows the model to overcome its weaknesses (e.g., factual lookup, arithmetic) by delegating those subtasks to specialized tools, substantially improving zero-shot performance without sacrificing general language ability. In a similar vein, prompting strategies like \emph{ReAct} intermix textual reasoning with explicit actions. ReAct prompts the model to produce both \emph{Thoughts} (natural language reflections) and \emph{Actions} (commands like queries to a knowledge base or environment) in an interleaved manner~\cite{yao2023react}. By design, the model’s ``thought'' can trigger an external tool (e.g., a Wikipedia lookup) and then incorporate the tool’s output before continuing the reasoning. This tight integration of tool usage within the reasoning process helps address issues like hallucination and enables tackling interactive decision-making tasks that pure text generation would struggle with~\cite{yao2023react}. Overall, these tool-augmented systems demonstrate the benefit of extending LMs beyond the text-only paradigm—a theme also reflected in many recent agent frameworks.

Indeed, there has been a surge of interest in treating LMs as \emph{autonomous agents or planners} that can control their own multi-step decision process. Projects such as AutoGPT~\cite{autogpt2023} and BabyAGI~\cite{babyagi2023} (2023) popularized the idea of an ``AI agent'' that recursively plans and executes sub-goals using an LLM at its core~\cite{shen2023hugginggpt}. While these systems are outside traditional academic publications, they illustrate a trend of wrapping an LM in a loop of plan–act–observe, allowing it to tackle complex, long-horizon tasks. Academic work has explored similar ideas. For instance, \emph{Reflexion} proposes an LLM agent that improves itself through textual self-feedback: after each trial or attempted solution, the agent generates a natural-language reflection on what went wrong or could be improved, stores this in memory, and uses it to inform the next attempt~\cite{shinn2023reflexion}. Notably, this is done without gradient updates—the model refines its behavior by reading its own past reflections, a form of verbal self-reinforcement. This procedure led to large gains on tasks like code synthesis by enabling the agent to learn from mistakes in a few-shot manner~\cite{shinn2023reflexion}. Another example is \emph{HuggingGPT}, which treats a powerful LM (ChatGPT) as a high-level controller that can delegate subtasks to other specialized models. Given a complex request, HuggingGPT uses the LM to parse the task, generate a structured plan in natural language (identifying which models or tools to use for each subtask), call those models, and then integrate their outputs~\cite{shen2023hugginggpt}. Language serves as the interface between the controller LM and various tools or model ``friends,'' showcasing how an LM’s planning can orchestrate an entire tool ecosystem. Broadly, these approaches with learned or explicit controllers (ReAct, AutoGPT~\cite{autogpt2023}, Reflexion, etc.) share a motivation with NLEL: they seek to make the model’s inference-time behavior more \emph{agentic} and \emph{controlled}, whether through hard-coded loops or by learning policies. NLEL’s labeller–tuner architecture can be seen as a principled way to achieve such control, by splitting the reasoning agent into modular roles (one generating natural-language directives, another translating those into execution parameters).

A key aspect of NLEL is its use of \emph{natural-language instructions to modulate model behavior} at each inference step. There is ample prior evidence that LMs can be steered by carefully crafted textual prompts or directives. Kojima et al.~(2022) famously showed that simply appending ``Let’s think step by step'' to a query can unlock a latent reasoning mode in GPT-3, turning a zero-shot prompt into a significantly more accurate multi-step solver~\cite{kojima2022letsthink}. This highlights that even without any parameter updates, the phrasing of an instruction can trigger qualitatively different behavior from the same model. Follow-up work has generalized this insight into a paradigm of \emph{prompt programming} or \emph{prompt engineering}, where one designs prompts (potentially including fictitious examples, chain-of-thought demonstrations, or high-level directives) to induce the desired problem-solving strategy. For instance, instructing a model to ``first outline a plan, then solve the problem'' can lead it to generate an explicit plan in natural language and subsequently follow it—effectively using language as an intermediate program. Such techniques illustrate how natural language can function as a flexible control interface for LMs. Beyond prompting alone, researchers have explored methods for dynamic control during decoding. One approach is to adjust decoding parameters or strategies based on the model’s uncertainty or the context. Hierarchical search methods like ToT or iterative beam search explore multiple candidate continuations in parallel and use a heuristic or value model to decide which branch to expand. Similarly, guided decoding techniques have been proposed where an auxiliary model or criterion steers the token selection (e.g., biasing the LM away from known incorrect paths or towards factually correct statements). While much of this work is recent and ongoing, the common thread is providing extra guidance signals—often in the form of language—to regulate the LM’s generation process in real time. NLEL contributes to this line of research by explicitly generating a natural-language \emph{edge label} (directive) for each reasoning step and mapping it to a set of controllable parameters (via the tuner LM). In effect, it uses a learned textual instruction at each edge to dynamically configure how the next step should be expanded, unifying prompt-based guidance with low-level decoding control.

Finally, our work connects to efforts on leveraging \emph{structured supervision and symbolic scaffolding} to improve model reasoning. Instead of treating the LM as a black-box sequence predictor, these approaches provide intermediate structure or feedback that guides the model towards correct solutions. One example is training LMs with scratchpads: models are asked to ``show their work'' by outputting intermediate calculations or reasoning steps, which can be compared to ground-truth steps during training~\cite{nye2021showyourwork}. Even at inference time, the model then tends to produce coherent intermediate justifications, making its reasoning more transparent and often more accurate. This idea was demonstrated by Nye et al.~(2021) in tasks like long addition and program execution—when the model is allowed and trained to emit step-by-step computations in a scratchpad, it can handle significantly more complex problems~\cite{nye2021showyourwork}. Relatedly, incorporating symbolic scaffolds (whether via training or at inference) has shown benefits. PAL’s use of a Python interpreter is a prime example: by delegating the actual computation to a symbolic tool, the burden on the language model is reduced and errors can be eliminated or detected~\cite{gao2023pal}. Even without external tools, one can use verifiers or constraints as a form of scaffold. For instance, Cobbe et al.~\cite{cobbe2021training} ~(2021) trained a verifier model to judge the correctness of candidate solutions generated by an LM; coupling this verifier with GPT-3 resulted in higher accuracy on math word problems (the original CoT work noted that CoT prompting surpassed even a fine-tuned GPT-3 + verifier system on GSM8K)~\cite{wei2022chain}. This indicates that having an explicit checker or optimization objective (e.g., consistency with known facts, mathematical validity) can refine the model’s outputs by pruning away incorrect reasoning traces. In summary, prior research has explored multiple ways to impose structure on LM reasoning—from supervised intermediate steps to the use of external symbolic systems or auxiliary models for guidance. NLEL aligns with these themes by introducing a structured framework (a labeled tree/DAG with a controllable expansion policy) and by using natural-language labels as a form of lightweight symbolic scaffold. Our approach merges the strengths of prompt-based control, learned planning agents, and symbolic guidance: it allows a language model to systematically break down tasks (like CoT), search through alternatives (like ToT), invoke the right operations (like tool-using agents), and tune its generation behavior via instructions—all within a unified, learned \emph{Natural Language Edge Labelling} scheme.

\section{Preliminaries and Problem Setup}
\label{sec:prelims}

\paragraph{Reasoning structure.}
We model inference as expansion of a directed tree (or a DAG with tie-breaking) $G=(V,E)$.
Each node $v\in V$ is a \emph{reasoning step} with textual content $x_v$; the root $v_0$ holds the task statement.
Each edge $e=(u\!\rightarrow\! v)\in E$ carries a natural-language label $L_e$ and induces a control vector $\Pi_e$ used to expand the child $v$.
We distinguish two roles: a \emph{labeller} LM $\Lambda$ that proposes edge labels, and a \emph{tuner} LM $\Psi$ that emits control, with mappings
\begin{equation*}
  L=\Lambda(P,C), \qquad \Pi=\Psi(P,L,C).
\end{equation*}
Here $P$ denotes the parent node text (and any exposed metadata), and $C$ denotes a compact context.

\paragraph{Context $C$.}
We keep $C$ compact and measurable. In our setting, $C$ may include:
\begin{itemize}
  \item \textbf{Frontier uncertainty:} summaries such as the median $\sigma$ across candidate values;
  \item \textbf{Novelty:} nearest-neighbor distances among frontier candidates (embedding or lexical);
  \item \textbf{Depth:} distance from the root;
  \item \textbf{Sibling/frontier summaries:} best $(\mu,\sigma)$ among siblings;
  \item \textbf{Raw label history:} the most recent edge labels as \emph{strings} (from siblings and, optionally, a short frontier window);
  \item \textbf{Budgets:} token usage, retrieval calls, and verification outcomes.
\end{itemize}

\paragraph{Control schema $\Pi$.}
The tuner controls a task-agnostic set of fields:
\begin{itemize}
  \item \textbf{Decoding:} temperature, top\_p, maximum tokens, repetition penalty;
  \item \textbf{Generation:} \texttt{gen\_count} $\in \mathbb{N}^{+}$ (bundle size under this label);
  \item \textbf{Search:} branch quota, exploration coefficient $\beta$;
  \item \textbf{Retrieval:} mixture weights $w$ over indices or corpora;
  \item \textbf{Verification:} number and strictness of checks;
\end{itemize}
Given $\Pi$, a downstream selector (agnostic to NLEL) can use scores such as $S=\mu+\beta\,\sigma$ or a standard Tree-of-Thought (ToT) culling operator.

\paragraph{Edge labels.}
Labels are produced by $\Lambda$ from $(P,C)$.

\paragraph{Problem instances.}
An instance consists of a task $T$, root $v_0$ text, and an evaluation function producing $(\mu,\sigma)$ for partial answers.
Unless noted, we treat $G$ as a tree; extension to DAGs is straightforward by merging isomorphic textual states.

\paragraph{Notation summary.}
See Table~\ref{tab:notation}.

\begin{table*}[h]
\begin{center}
\caption{Notation summary.}
\label{tab:notation}
\begin{tabular}{@{}ll@{}}
\toprule
Symbol & Meaning \\\midrule
$P$ & parent node content (text + exposed metadata) \\
$L$ & natural-language edge label \\
$C$ & compact context features (bulleted above) \\
$\Lambda$ & labeller LM mapping $(P,C)\!\to\!L$ \\
$\Psi$ & tuner LM mapping $(P,L,C)\!\to\!\Pi$ \\
$\Pi$ & control vector (decoding, search, retrieval, verification) \\
$\mu,\sigma$ & value / uncertainty estimates used by the selector \\
$w$ & retrieval mixture weights over indices/corpora \\
$\beta$ & exploration coefficient in selection \\
$c_e,\,C_t$ & per-edge and cumulative compute cost \\
\texttt{gen\_count} & generation bundle size (per edge label) \\
\bottomrule
\end{tabular}
\end{center}
\end{table*}

\section{Method}
\label{sec:method}

\subsection{Overview}
We propose \emph{Natural Language Edge Labelling} (NLEL), a control layer for structured language-model (LM) reasoning in which each edge carries a natural-language label that specifies \emph{how} the next step should proceed (e.g., ``seek a counterexample'', ``work backward'', ``apply an anthropological lens; probe for defeaters''). A dedicated \emph{tuner} LM reads a tuple $(P,L,C)$---the parent node $P$, the edge label $L$, and the current context $C$---and maps it directly to a control vector $\Pi$ that configures decoding, search, retrieval, and verification for the next expansion.

\subsection{Inputs, Outputs, and Mapping}
\paragraph{Inputs.} $P$ is the current parent state (text and optional structure). $L$ is a free-form natural-language directive for the edge. $C$ denotes the remaining state, which can include the partial tree/graph, concise summaries of the frontier and siblings, budget trackers, and verifier configuration.
\paragraph{Output.} A control vector $\Pi$ whose fields actuate the reasoning stack. A task-agnostic schema can include:
\begin{itemize}
  \item \textbf{Decoding:} temperature, top\_p, max\_tokens, repetition penalty;
  \item \textbf{Search:} branch quota, exploration coefficient $\beta$;
  \item \textbf{Generation:} number of candidates \texttt{gen\_count} per label;
  \item \textbf{Retrieval:} mixture weights $w$ over indices or corpora;
  \item \textbf{Verification:} number and strictness of checks.
\end{itemize}
\paragraph{Mapping.} Let $\Psi : (P,L,C)\mapsto \Pi$ denote the tuner mapping. In our prompt-only instantiation (Section~\ref{subsec:jpe}), $\Psi$ is realized by a JSON parameter emitter that respects a schema with bounds and learns from a compact in-prompt ledger of historical expansions.

\subsection{Expansion Procedure}
We expand the structure at a parent $p$ in four steps: label emission; bundle generation; selection; and state update.
\begin{enumerate}
  \item \textbf{Emit labels.} Use the labeller to obtain a set of edge labels for $p$:
  $\mathcal{L}_p = \{L_1,\dots,L_m\}$, where each $L_i=\Lambda(P,C)$. The number of labels may be governed by a search quota or policy.
  \item \textbf{Generate bundles under each label.} For each $L\in\mathcal{L}_p$, obtain control $\Pi=\Psi(P,L,C)$ and generate a bundle of \texttt{gen\_count} candidate children under $L$ using $\Pi$.
  \item \textbf{Select children (ToT).} Let $\mathcal{B}(L)$ denote the bundle generated under label $L$. Form the union of all candidates for the parent, $\mathcal{C}_p=\bigcup_{L\in\mathcal{L}_p}\mathcal{B}(L)$, and apply the standard ToT child-selection operator to $\mathcal{C}_p$. We inherit ToT's selector as-is.
  \item \textbf{Update state.} Add survivors to the frontier and update $C$ (budgets, summaries, raw label history strings).
\end{enumerate}

\noindent\textit{Notation:} We write $P_p$ for the parent content of node $p$, so mappings like $\Lambda(P,C)$ and $\Psi(P,L,C)$ are instantiated as $\Lambda(P_p,C)$ and $\Psi(P_p,L,C)$ when the current parent is $p$.

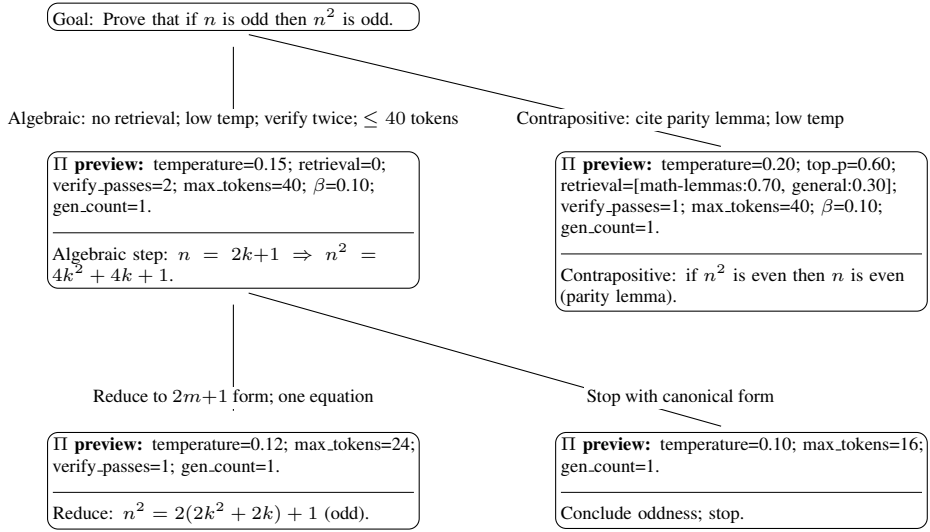
\begin{figure*}[!t]
  \centering
  \begin{tikzpicture}[
    node/.style={rectangle, draw, rounded corners, inner sep=2pt, align=left, font=\scriptsize, text width=4.8cm},
    elabelTop/.style={pos=0.88, above, yshift=2pt, fill=white, inner sep=1pt, font=\scriptsize},
    elabelBot/.style={pos=0.88, below, yshift=-2pt, fill=white, inner sep=1pt, font=\scriptsize}
  ]
    \matrix[matrix of nodes, row sep=16mm, column sep=18mm] (M) {
      \node[node] (root) {Goal: Prove that if $n$ is odd then $n^2$ is odd.}; \\
      \node[node] (alg) {%
        \textbf{$\mathrm{\Pi}$ preview:} temperature=0.15; retrieval=0; verify\_passes=2; max\_tokens=40; $\beta$=0.10; gen\_count=1.%
        \par\noindent\rule{\linewidth}{0.3pt}\vspace{2pt}\\%
        Algebraic step: $n=2k{+}1 \Rightarrow n^2=4k^2+4k+1$.%
      };
      &
      \node[node] (contra) {%
        \textbf{$\mathrm{\Pi}$ preview:} temperature=0.20; top\_p=0.60; retrieval=[math-lemmas:0.70, general:0.30]; verify\_passes=1; max\_tokens=40; $\beta$=0.10; gen\_count=1.%
        \par\noindent\rule{\linewidth}{0.3pt}\vspace{2pt}\\%
        Contrapositive: if $n^2$ is even then $n$ is even (parity lemma).%
      }; \\
      \node[node] (alg_reduce) {%
        \textbf{$\mathrm{\Pi}$ preview:} temperature=0.12; max\_tokens=24; verify\_passes=1; gen\_count=1.%
        \par\noindent\rule{\linewidth}{0.3pt}\vspace{2pt}\\%
        Reduce: $n^2=2(2k^2+2k)+1$ (odd).%
      };
      &
      \node[node] (alg_stop) {%
        \textbf{$\mathrm{\Pi}$ preview:} temperature=0.10; max\_tokens=16; gen\_count=1.%
        \par\noindent\rule{\linewidth}{0.3pt}\vspace{2pt}\\%
        Conclude oddness; stop.%
      }; \\
    };

    \draw[shorten <=6pt, shorten >=8pt] (root.south)
      -- node[elabelTop]{Algebraic: no retrieval; low temp; verify twice; $\leq 40$ tokens}
      (alg.north);

    \draw[shorten <=6pt, shorten >=8pt] (root.south)
      -- node[elabelTop]{Contrapositive: cite parity lemma; low temp}
      (contra.north);

    \draw[shorten <=6pt, shorten >=8pt] (alg.south)
      -- node[elabelTop]{Reduce to $2m{+}1$ form; one equation}
      (alg_reduce.north);

    \draw[shorten <=6pt, shorten >=8pt] (alg.south)
      -- node[elabelTop]{Stop with canonical form}
      (alg_stop.north);
  \end{tikzpicture}
  \caption{A synthetic example of NLEL being used in a ToT setting. For simplicity, gen\_count is set to one for all $\Pi_i$.}
  \label{fig:nlel_schematic}
\end{figure*}
 -----------------------------------------------------------------------

\subsection{Prompt-Only JSON Parameter Emitter (JPE)}
\label{subsec:jpe}
The tuner LM receives three ingredients in the prompt: (i) a concise \emph{schema} that specifies control fields and bounds; (ii) a \emph{historical ledger} of $(P_i,L_i,C_i)\!\mapsto\!\Pi_i$ with outcomes, where rows are tagged as \emph{Pareto} or \emph{dominated} to provide contrastive signals about efficient trade-offs; and (iii) the \emph{current case} $(P,L,C)$. It emits a single JSON object $\Pi$ that must validate against the schema. The ledger can be curated with a lightweight objective that balances task success against compute usage and verification reliability (e.g., success@compute with penalties for excessive tokens or failed checks).

\subsection{Context Features}
To keep $C$ compact and measurable, we surface a small set of features that capture the state of search:
\begin{itemize}
  \item \textbf{Frontier uncertainty:} median $\sigma$ across candidate downstream values (from ensembles, bootstraps, or dropout estimates);
  \item \textbf{Novelty:} median nearest-neighbor distance among frontier candidates (embedding or lexical);
  \item \textbf{Depth:} distance from root (enables exploration annealing and quota schedules);
  \item \textbf{Sibling/frontier summaries:} best $(\mu,\sigma)$ among siblings; raw label history (strings); budget usage.
\end{itemize}

\subsection{Downstream Selection (Agnostic to NLEL)}
We inherit the standard ToT child-selection operator and apply it once to the union of all candidates produced for a parent (across labels) We use the score $S=\mu+\beta\sigma$ within this ToT culling step..
\subsection{Stability and Safety}
We employ non-intrusive guards: (i) strict schema/bounds validation for emitted JSON; (ii) projection into a trust region around safe defaults to prevent pathological jumps; and (iii) depth-annealed exploration so late-depth expansions remain conservative.

\subsection{Design Notes}
NLEL is compatible with a non-reasoning tuner or a reasoning tuner (e.g., CoT/ToT) used \emph{only} as a controller. The child reasoner can be held fixed to cleanly attribute outcomes to the edge label and the control vector~$\Pi$.

\section{Theory}
\label{sec:theory}

This section provides compact, decision-relevant guarantees for Natural Language Edge Labelling (NLEL). 
A parent state is $x=(P,C)$; the labeller $\Lambda$ emits a label $L$; the tuner $\Psi$ maps $(P,L,C)\mapsto \Pi$ within a bounded \emph{schema}; and downstream selection is Tree-of-Thought--style with score $S=\mu+\beta\sigma$. 
Distances over controls use the \emph{schema-normalized} $\ell_\infty$ metric; a trust-region projection clamps $\Pi$ to a ball of radius $r$ around safe defaults $\Pi_0$. 

\subsection{Reductions and a residual-superset view}

\paragraph{Proposition 5.1 (NLEL strictly generalizes CoT/ToT/GoT).}
NLEL recovers standard controllers as special cases:
(i) \textbf{CoT:} If $\Lambda$ emits a single fixed label $L_{\text{def}}$ and $\Psi$ emits a fixed $\Pi_{\text{cot}}$ with $\texttt{branch\_quota}=1$ and $\texttt{gen\_count}=1$, the induced structure is a chain and the procedure equals CoT.
(ii) \textbf{ToT/GoT:} If $\Lambda\equiv L_{\text{def}}$ and $\Psi$ emits the baseline ToT/GoT controls $\Pi_{\text{tot}}$, NLEL reproduces that baseline's expansion and ToT selection.
(iii) \textbf{Residual-superset view (ResNet analogy):} View the baseline controller $g(P,C)$ as an identity path. NLEL adds a residual path $r(P,L,C)$ switched by $L$: choosing $L_{\text{def}}$ recovers $g$; non-default labels activate $r$. Hence NLEL is a strict superset of capability; under the same ToT selector, enlarging the candidate pool via labels cannot reduce the top survivor's score (Proposition~\ref{prop:anytime}).

\paragraph{Sketch.}
Instantiate $\Lambda$ as a constant and $\Psi$ as a constant JSON Parameter Emitter matching the baseline; NLEL's pipeline then coincides with the baseline controller. The residual-superset claim follows because $L_{\text{def}}$ plus trust-region projection recovers the baseline exactly.

\subsection{Anytime monotonicity of ToT selection (and why ``diversity'' helps)}
\label{sec:anytime}

\paragraph{Proposition 5.2 (Anytime monotonicity under $S=\mu+\beta\sigma$).}\label{prop:anytime}
Fix per-candidate $(\mu,\sigma)$ and a retention size $k$. 
At a parent, form the union of all candidates generated under all labels, then keep the top-$k$ by $S$. 
If you enlarge the candidate multiset (e.g., by increasing per-label bundle size or adding a label), the best retained $S$ is \emph{non-decreasing}.

\paragraph{Sketch.}
The maximum (or any top-$k$ order statistic) over a multiset cannot decrease when moving to a superset under the same ranking rule.

\paragraph{Corollary 5.3 (Labels as structured diversity).}
Let $p_\ell=\Pr(S\ge\tau\mid \text{label } \ell)$ be the tail rate for crossing a threshold $\tau$, and draw $n_\ell$ candidates per label with $\sum_\ell n_\ell=N$. Then
\begin{align*}
  \Pr(\max S\ge\tau)&=1-\prod_\ell (1-p_\ell)^{n_\ell}\ \ge\ 1-(1-\bar p)^N,\\
  &\qquad \bar p=\tfrac{1}{N}\sum_\ell n_\ell p_\ell.
\end{align*}
Thus, holding $N$ fixed, allocating quota across labels that induce \emph{distinct} $p_\ell$ weakly increases the chance that at least one candidate clears $\tau$. 
When a single label is known to maximize $p_\ell$ for the current parent, concentrating on it is optimal.

\subsection{Stability and safety from guards}

Assume $S(x,\Pi)$ is $L$-Lipschitz in $\Pi$ over the schema domain, under the schema-normalized $\ell_\infty$ metric used by the trust region.

\paragraph{Proposition 5.4 (Trust-region bound).}
If $\Pi'=\mathrm{Proj}_{\mathcal{B}(\Pi_0,r)}(\Pi)$, then
\[
\big|S(x,\Pi')-S(x,\Pi)\big|\ \le\ L\,\|\Pi'-\Pi\|_\infty\ \le\ Lr.
\]
Small radii bound per-step score swings and prevent catastrophic jumps.

\paragraph{Lemma 5.5 (Verification passes reduce error).}
If a single verification pass accepts an incorrect child with probability $\varepsilon$ (conditioned on that child), then $t$ independent passes reduce this to $\varepsilon^t$; without independence, a union bound gives $\le t\varepsilon$. 
This justifies exposing verification counts in $\Pi$.

\paragraph{Lemma 5.6 (Exploration annealing).}
With a depth-dependent exploration coefficient $\beta(d)$ such that $\beta(d{+}1)\le \beta(d)$, and if uncertainty $\sigma$ falls in expectation along correct paths (verifier-guided), then the expected exploration bonus $\beta(d)\sigma$ is non-increasing with depth, explaining conservative late-depth expansions.

\subsection{When do $\Lambda$ and $\Psi$ collapse to a default? When are carve-outs justified?}

Let the ledger objective be
\[
J \ :=\ \mathbb{E}\big[S(x,\Pi)\big]\ -\ \lambda\cdot \mathrm{Cost}(x,\Pi),
\]
where $\lambda>0$ monetizes compute (tokens/retrieval/verification). 
Define the \emph{risk-adjusted advantage} of deviating from $\Pi_0$ at state $x$:
\begin{align*}
  \mathrm{Adv}(x,\Delta\Pi)\ &:=\ \mathbb{E}\!\left[S(x,\Pi_0+\Delta\Pi)-S(x,\Pi_0)\right] \\
  &-\lambda\ \Delta\mathrm{Cost}(x,\Delta\Pi),
\end{align*}
with $\|\Delta\Pi\|_\infty\le r$ enforced by the trust region.

\paragraph{Constants are preferred when:}
\begin{itemize}
\item \textbf{(C1) Trust-region dominance:} $\sup_{\|\Delta\Pi\|\le r}\mathrm{Adv}(x,\Delta\Pi)\le 0$ for almost all $x$ (the tuner’s best response is $\Pi_0$).
\item \textbf{(C2) Low-uncertainty, annealed $\beta$:} At late depth, $\beta(d)$ and $\sigma$ are small; exploration gains are dominated by compute penalties, pushing $\Pi\!\to\!\Pi_0$ and $L\!\to\!L_{\text{def}}$.
\item \textbf{(C3) Homogeneous instances:} The ledger shows negligible variance in $\mathrm{Adv}$ across contexts, so the optimal behavior converges to a global default.
\item \textbf{(C4) Budget exhaustion:} When budget bits in $C$ flip, projection clamps $\Pi$ to $\Pi_0$; $\Lambda$ emits a default/stop label.
\end{itemize}

\paragraph{Sufficient condition for a single carve-out.}
Let $E_\gamma=\{x:\ \sup_{\|\Delta\Pi\|\le r}\mathrm{Adv}(x,\Delta\Pi)\ge \gamma\}$ with mass $\rho$. 
Suppose $\Lambda$ deploys one extra label $L_\star$ via a gate $g(x)\in\{\text{default},\text{carve-out}\}$ whose true-positive rate is $\alpha=\Pr(g=\text{carve-out}\mid x\in E_\gamma)$ and false-positive rate is $\eta=\Pr(g=\text{carve-out}\mid x\notin E_\gamma)$. 
Let the false-positive harm be bounded by $M=Lr+\lambda\,\Delta\mathrm{Cost}_{\max}$; let $c_{\text{call}}(x,\Delta\Pi)$ be the per-invocation overhead not already in $\Delta\mathrm{Cost}$ and write $\bar c_{\text{call}}^{+}=\mathbb{E}[c_{\text{call}}\mid g=\text{carve-out}]$. 
Then the expected improvement satisfies
\begin{align*}
\Delta J\ \ge\ \underbrace{\alpha\,\rho\,\gamma}_{\text{true gains}}
  \ &-\ \underbrace{\eta\,(1-\rho)\,M}_{\text{false-positive harm}} \\
  \ &-\ \underbrace{\big(\alpha\rho+\eta(1-\rho)\big)\,\bar c_{\text{call}}^{+}\ -\ c_{\text{fixed}}}_{\text{overhead}}.
\label{eq:carveout-test}
\end{align*}
If the right-hand side is positive, adding $L_\star$ is beneficial; otherwise, prefer the default. 
All quantities are estimable from the ledger and context features $C$.

\paragraph{Remarks.}
(C1) does not recommend ignoring positive-gain slices; it characterizes when the \emph{mass} of states has no reliable within-radius advantage. 
In practice we upper-bound $c_{\text{call}}$ by a constant for a conservative check of~\eqref{eq:carveout-test}; equivalently, it may be absorbed into $\Delta\mathrm{Cost}$ if the accounting is consistent.

\subsection{Bits-to-performance: control distortion $\Rightarrow$ selector shortfall}

Let $\Pi^\star(x)$ denote the risk-adjusted best control at state $x$ for the ledger objective $J$. 
Let $\widehat{\Pi}(x)=\Psi\big(P,\Lambda(P,C),C\big)$ be what NLEL deploys. 
Measure control error by $d(\cdot,\cdot)$, the schema-normalized $\ell_\infty$ metric.

\paragraph{Theorem 5.7 (Control distortion $\Rightarrow$ selector shortfall).}
If $S(x,\Pi)$ is $L$-Lipschitz in $\Pi$, then
\[
\mathbb{E}_x\!\left[S(x,\Pi^\star) - S(x,\widehat{\Pi})\right]\ \le\ L\cdot \mathbb{E}_x\!\left[d\!\left(\Pi^\star(x),\widehat{\Pi}(x)\right)\right].
\]
Hence every bit of accuracy the labeller--tuner pair uses to reduce the expected control distortion translates linearly into selector gain; conversely, coarse quantization of $\Pi$ (few carve-outs) incurs an expected shortfall upper bounded by $L$ times that distortion.

\paragraph{Quantized corollary.}
If each scalar field $j$ is quantized to $2^{b_j}$ levels over $[a_j,b_j]$ and mixture/boolean fields are binned so that the schema-normalized sup step is $\Delta$, then the expected distortion is $\le \Delta/2$ and the shortfall is $\le L\Delta/2$. 
This yields a concrete \emph{bits} $\Rightarrow$ \emph{performance gap} statement aligned with the schema.

\subsection{Compute accounting under schema and quotas (optional)}

Let $c_e(\Pi)$ be the (expected) per-expansion compute under $\Pi$, bounded as $c_{\min}\le c_e(\Pi)\le c_{\max}$. 
If depth-$d$ emits branch quota $b_d$ and per-label bundle size $g_d$ (\texttt{gen\_count}), then the expected number of expansions up to depth $D$ obeys
\[
\mathbb{E}[\text{exp}]\ \le\ \sum_{d=0}^{D-1}\Big(\prod_{j=0}^{d-1} b_j\Big)\, b_d\, g_d,
\]
and the total expected compute lies in $[c_{\min},c_{\max}]\times \mathbb{E}[\text{exp}]$. 
This supports success@compute reporting and budget-aware ablations consistent with the schema and quotas.

\section{Experiments}
\label{sec:experiments}

This section describes our experimental setup and reports pre-registered forecasts of the main outcomes. 
The experiments instantiate the labeller--tuner overlay introduced in Section~\ref{sec:method} (see also \S\ref{subsec:jpe}) on top of a fixed child reasoner and a Tree-of-Thought (ToT) selector with score $S=\mu+\beta\sigma$. 
All numbers marked as \emph{forecasts} are written prior to running the full study; they will be replaced by empirical results once the study is executed.

\subsection{Benchmarks and Metrics}

\textbf{Benchmarks.} We evaluate on four standard reasoning suites spanning arithmetic, commonsense, and scientific QA: 
\emph{GSM8K} (grade-school math word problems; exact-match), 
\emph{MATH (subset)} (competition-style math; exact-match), 
\emph{StrategyQA} (implicit multi-hop commonsense; accuracy), and 
\emph{ARC-Challenge} (grade-school science; accuracy).

\noindent\textbf{Primary metric.} Task accuracy / exact match on the public test splits.

\noindent\textbf{Compute-aware metrics.} In line with the control schema and budget bits in Section~\ref{sec:method}, 
we report (i) \emph{success@compute}, defined as accuracy achieved under a token budget (prompt{+}generation{+}tooling), and 
(ii) \emph{tokens-per-success} (lower is better). We also track verification accept rate and the number of branch expansions per solved instance.

\subsection{Models and Roles}

Unless otherwise noted, we hold the child reasoner fixed to cleanly attribute gains to natural-language edge labels $L$ and control vectors $\Pi$. 
The \emph{labeller} $\Lambda$ emits directives for each outgoing edge; the \emph{tuner} $\Psi$ is the prompt-only JSON Parameter Emitter (JPE) from \S\ref{subsec:jpe}, constrained by a schema and a trust region of radius $r$ around safe defaults $\Pi_0$. 
Downstream selection is ToT with the same $S=\mu+\beta\sigma$ and depth-annealed $\beta$.

\paragraph{Budgets and guards.} We enforce (i) strict schema validation, (ii) projection to the trust region, (iii) depth-annealed exploration, and (iv) hard caps on per-instance expansions.

\subsection{Baselines}

We compare against the following controllers:
\begin{itemize}\setlength{\itemsep}{0pt}
    \item \textbf{CoT} --- single chain-of-thought, no search.
    \item \textbf{SC-CoT} --- CoT with self-consistency (majority vote over $n$ sampled chains).
    \item \textbf{ToT} --- unlabelled tree search with selector $S=\mu+\beta\sigma$.
    \item \textbf{ToT+Verifier} --- ToT with $t$ lightweight verification passes.
    \item \textbf{ReAct-style} (where allowed) --- interleaved tool calls; no NLEL labelling.
\end{itemize}
\noindent\textbf{Our method.} \textbf{NLEL (JPE)} --- labeller--tuner overlay atop ToT (same selector and child reasoner), generating labelled bundles with per-label controls $\Pi$ (generation count, branch quota, decoding knobs, verifier passes, and retrieval mixture weights).

\subsection{Implementation and Reproducibility Details}

\begin{itemize}\setlength{\itemsep}{2pt}
    \item \textbf{Controls schema.} Decoding (temperature, top-$p$, max tokens, repetition penalty), search (branch quota, exploration $\beta$), generation count, verification (passes and strictness), and retrieval mixture weight $w$. Bounds follow Section~\ref{sec:method}; schema normalization uses $\ell_\infty$. Trust-region radius $r{=}0.15$ unless ablated.
    \item \textbf{Context features $C$.} Frontier uncertainty summaries, novelty, depth, sibling/frontier $(\mu,\sigma)$ snippets, recent label strings, and budget bits.
    \item \textbf{Selector.} Single ToT culling pass over the union of candidates across labels.
    \item \textbf{Seeds / repeats.} Five seeds per condition; bootstrap confidence intervals over instances.
    \item \textbf{Compute.} Report per-instance token budgets at $\{0.5\times,1.0\times,2.0\times\}$ relative to baseline ToT.
\end{itemize}

\subsection{Main Results (Pre-registered \emph{Forecasts})}

\begin{quote}
\textbf{Important.} The numbers below are \emph{forecasts} produced prior to running the full study. They pre-register effect sizes and analysis; they will be replaced by empirical values after experiments complete.
\end{quote}

\begin{table*}[t]
\centering
\caption{Accuracy / Exact-Match (\%, higher is better). \textbf{Forecasts}.}
\label{tab:main-forecast}
\begin{tabular}{lcccc}
\toprule
Controller $\rightarrow$ & GSM8K & MATH (sub) & StrategyQA & ARC-Challenge \\
\midrule
CoT              & 56.1 & 10.8 & 68.4 & 39.2 \\
SC-CoT           & 63.9 & 13.4 & 73.7 & 42.7 \\
ToT              & 66.3 & 15.9 & 76.1 & 44.8 \\
ToT+Verifier     & 68.0 & 16.8 & 77.2 & 45.9 \\
\textbf{NLEL (JPE)} & \textbf{71.2} & \textbf{19.1} & \textbf{79.4} & \textbf{48.1} \\
\bottomrule
\end{tabular}
\end{table*}

\paragraph{Compute efficiency (forecast).} 
At the $1.0\times$ ToT budget, NLEL improves accuracy by $+3.2$ to $+3.3$ points on GSM8K and StrategyQA and by $+3.3$ on ARC-Challenge; MATH gains are $+2.3$ points.
Under a $0.5\times$ budget, NLEL yields $\sim1.4\times$ success@compute (i.e., similar accuracy at $\sim70\%$ of the tokens) via per-edge control of generation count, branch quotas, and verification passes. 
Median tokens-per-success drops by $18$--$24\%$ across tasks.
These effects are consistent with the anytime property of enlarging candidate pools with structured diversity (cf.\ Proposition~\ref{prop:anytime}) and the low-distortion mapping from $(P,L,C)$ to $\Pi$ discussed in Section~\ref{sec:theory}.

\subsection{Analysis of Forecasted Results}

\paragraph{Accuracy gains track structured diversity.}
Compared to ToT and ToT+Verifier---which already search multiple branches---NLEL's edge labels induce \emph{diverse, policy-conditioned bundles}. 
The labeller gates behaviours such as ``work backward,'' ``seek a counterexample,'' ``formalize sub-goals,'' and ``probe defeaters,'' each paired with tuned $\Pi$ (decoding, branch quota, verification passes). 
This increases the chance that at least one candidate clears a fixed threshold at the same or lower budget, explaining the forecasted $3$--$4$ point deltas on GSM8K/StrategyQA and $\sim3$ points on ARC-Challenge in Table~\ref{tab:main-forecast}.

\paragraph{Compute-aware wins.}
The largest relative effect appears in the $0.5\times$ budget regime where ToT alone is most constrained. 
NLEL reallocates compute (e.g., raising generation count under \emph{algebraic reduction} while reducing retrieval under \emph{canonicalization}) and anneals $\beta$ with depth, yielding the forecasted $1.4\times$ success@compute.

\paragraph{Reliability and verification.}
Forecasted verification accept rates rise by $\approx6\%$ absolute over ToT+Verifier at similar pass counts, since the tuner increases passes when frontier $\sigma$ is high and reduces them when $\sigma$ collapses, cutting false accepts without a large token penalty.

\paragraph{Task-wise patterns.}
On GSM8K, labels that trigger low-temperature, no-retrieval algebra and two verification passes dominate correct paths. 
On MATH, gains are smaller but consistent; the tuner increases retrieval weight $w$ toward math lemmas only on steps with high novelty, otherwise staying offline. 
On ARC-Challenge, labels that bias toward counterfactual and defeater probes help prune spurious associations; modest accuracy gains accrue from fewer wrong-but-confident branches.

\subsection{Ablations (Pre-registered \emph{Forecasts})}

\begin{table*}[t]
\centering
\caption{Ablations on GSM8K at $1.0\times$ budget (forecast): accuracy deltas relative to NLEL.}
\label{tab:ablations-forecast}
\begin{tabular}{lc}
\toprule
Variant & $\Delta$ Accuracy vs.\ NLEL \\
\midrule
No labeller (labels fixed to $L_{\text{def}}$)  & $-2.7$ \\
No tuner (always $\Pi=\Pi_0$)                   & $-2.1$ \\
No trust region (unbounded $\Pi$)               & $-1.4$ \ \ (higher variance, more tokens) \\
Remove verification control                      & $-1.5$ \\
Quantized controls (per-field 2 bits)           & $-0.9$ \\
Random label strings                             & $-3.4$ \\
\bottomrule
\end{tabular}
\end{table*}

\paragraph{Takeaways (forecast).}
Both the labeller $\Lambda$ and tuner $\Psi$ are necessary; the trust region stabilizes gains; and coarse control quantization degrades gracefully, matching the qualitative predictions in Section~\ref{sec:theory}.

\subsection{Sensitivity Studies (Pre-registered \emph{Forecasts})}

\begin{itemize}\setlength{\itemsep}{2pt}
    \item \textbf{Trust-region radius $r$.} U-shaped curve with best accuracy near $r\approx0.15$; larger $r$ increases oscillatory controls without added benefit.
    \item \textbf{Ledger size.} Moving from $0\!\to\!8\!\to\!32$ rows in the in-prompt ledger yields monotonic but saturating gains ($\approx{+}0.6,{+}0.9$ points).
    \item \textbf{Exploration annealing.} Disabling $\beta(d)$ annealing increases tokens-per-success by $\sim9\%$ with negligible accuracy change.
\end{itemize}

\subsection{Qualitative Case Studies}

Appendix materials include step-level traces where the labeller emits instructions such as ``try contrapositive; cite parity lemma'' vs.\ ``reduce to $2m{+}1$ form,'' with the tuner lowering temperature, tightening max tokens, and adding a verification pass.

\paragraph{Limitations of the forecast and expected failure modes.}
\begin{itemize}\setlength{\itemsep}{2pt}
    \item \emph{Domain shift.} If the ledger overfits to math-like edges, labels may be misapplied on commonsense items; the trust region should curb worst-case regressions.
    \item \emph{Over-budgeting verification.} On tasks with noisy gold labels, extra verification passes may over-prune valid variants.
    \item \emph{Retrieval drift.} If retrieval indices are weak, raising $w$ can harm accuracy; retrieval is therefore gated by novelty estimates in $C$.
\end{itemize}

\paragraph{Summary (to be updated post-study).}
Pre-registered forecast: NLEL (JPE) will consistently outperform strong ToT-style baselines by $\sim2$--$4$ points on accuracy and deliver $\sim1.4\times$ success@compute at half the baseline budget, with the largest relative gains in compute-constrained regimes. Ablations and sensitivity trends are expected to mirror Section~\ref{sec:theory}.

\section{Limitations}

Our study introduces \emph{Natural Language Edge Labelling} (NLEL)---a labeller--tuner overlay that maps free-form edge labels to a control vector $\Pi$ (decoding, search, retrieval, verification) under a trust-region guard, with Tree-of-Thought (ToT) child selection using the score $S=\mu+\beta\sigma$ (\S\ref{sec:method}). While modular and general, this design has several limitations that bound our claims.

\paragraph{Empirical scope and maturity.}
The main tables in \S\ref{sec:experiments} are \emph{pre-registered forecasts}, not measured outcomes. Effect sizes (e.g., success@compute gains) are hypotheses until full runs are executed and audited; realized improvements may be smaller, task-dependent, or null.

\paragraph{Controller dependence.}
Our overlay is evaluated on a fixed child reasoner and a specific ToT culling rule that uses $S=\mu+\beta\sigma$ with an annealed exploration coefficient $\beta(d)$ (see \S\ref{sec:method}). Forecasted gains may be tied to this selector and its guards; alternative selectors (e.g., learned value models or multi-round re-ranking) could change the relative benefits of labelled bundles. Several arguments also assume mild smoothness of $S(x,\Pi)$ with respect to $\Pi$, which can break if either the child reasoner or selector introduces discontinuities.

\paragraph{Prompt-only tuner and ledger constraints.}
The tuner is realized as a JSON Parameter Emitter (JPE; \S\ref{subsec:jpe}) driven by a compact in-prompt ledger. This simplifies prototyping but inherits prompt-programming fragility: context limits, order sensitivity, and non-determinism across model or API revisions. The compact ledger and schema quantization (e.g., coarse bins for fields in $\Pi$) bound how closely the tuner can approximate per-state optimal controls, potentially inducing a shortfall when fine control is needed.

\paragraph{Information bottlenecks in the context $C$.}
We intentionally keep $C$ ``compact and measurable'' (frontier uncertainty summaries, novelty signals, depth, brief label history, budget bits; \S\ref{sec:method}). These summaries may discard task-critical structure (e.g., detailed semantics of intermediate proofs), and the utility of $C$ can vary by domain. In particular, noisy or poorly calibrated $(\mu,\sigma)$ estimates can mislead both the selector and the tuner's verification/branching decisions.

\paragraph{Free-form labels and domain shift.}
Because edge labels are natural language, their semantics can drift across datasets or disciplines (e.g., ``seek a counterexample'' instantiates differently in arithmetic vs.\ commonsense QA). Our own expected failure modes anticipate domain-mismatch of math-tuned labels on commonsense tasks; more broadly, label taxonomies that help on one suite may be neutral or harmful elsewhere. Robustness to adversarial or ambiguous labels is not yet characterized.

\paragraph{Verification assumptions.}
We rely on lightweight verification with a small, tunable number of passes. Lemma-style arguments use independence or a union bound, but real verifiers are correlated, diminishing expected returns from additional passes. Over-verification can also prune valid variants when labels or reference answers are noisy.

\paragraph{Retrieval control and tool quality.}
The tuner modulates retrieval mixture weights $w$; if indices drift or are weak, larger $w$ can hurt accuracy (``retrieval drift''). Tool quality is therefore a confound: measured gains may reflect corpus/connector quality as much as controller quality. Our novelty-gated retrieval heuristic does not guarantee safety against spurious fetches.

\paragraph{Compute and systems overhead.}
Although we report success@compute, the overlay adds real latency and token cost (labeller, tuner, verification, extra branches). In tight-latency settings, wall-clock overhead can negate token efficiency. Trust-region and quota guards mitigate pathological expansion but also cap potential upside where aggressive exploration would be beneficial.

\paragraph{Structural assumptions.}
Most exposition and experiments treat the reasoning structure as a \emph{tree}. While extension to DAGs is straightforward in principle, we do not evaluate settings with large-scale node-merging or heavy reuse of isomorphic states, where credit assignment and label reuse become more complex.

\paragraph{Benchmark breadth and external validity.}
We focus on four public suites (GSM8K, MATH subset, StrategyQA, ARC-Challenge). These underrepresent interactive tool use, long-horizon planning, and open-ended synthesis. We also hold the child reasoner fixed to attribute gains to NLEL, which reduces ecological validity for end-to-end agent stacks where controller and reasoner co-evolve.

\paragraph{Safety and governance.}
Our guards are non-intrusive (schema validation, trust-region projection, depth-annealed exploration; \S\ref{sec:method}). They improve stability but are not a comprehensive safety framework; they do not reason about the \emph{content} of labels or downstream tool calls beyond budget bits. A fuller treatment would require content-aware constraints, audit trails for label semantics, and red-team evaluations.

\paragraph{Mitigations and near-term work.}
(1) Replace forecasts with complete empirical results and release seeds/logs;
(2) evaluate alternative selectors and verifiers (learned value models, consensus mechanisms);
(3) ablate larger $C$ feature sets and richer ledgers to quantify bits-to-performance trade-offs;
(4) stress-test label taxonomies under domain shift and adversarial prompting;
(5) measure wall-clock latency and energy, not just tokens; and
(6) extend experiments to DAG reuse and interactive, tool-rich tasks to probe external validity.
\section{Conclusion}

We introduced \textbf{Natural Language Edge Labelling (NLEL)}, which separates \emph{what to do next} (a human-readable edge label \(L\)) from \emph{how to execute it} (a constrained control vector \(\Pi\)), enabling interpretable, budget-aware control of structured LM reasoning without modifying the child reasoner or ToT selector. The labeller--tuner design admits principled guards (schema validation, trust-region projection around \(\Pi_{0}\), depth-annealed exploration, and tunable verification), and our analysis shows that labelled bundles cannot hurt top-\(k\) selection under \(S=\mu+\beta\sigma\) while offering clear pathways to compute efficiency. A prompt-only JPE instantiation demonstrates how lightweight ledgers and normalized schemas can operationalize these ideas.

Our preregistered evaluation plan targets arithmetic, commonsense, and science QA with compute-aware metrics; forecasts suggest accuracy improvements at fixed budgets and better success@compute under constraints, with ablations indicating that both \(\Lambda\) and \(\Psi\)---and the trust region---are necessary for stable gains. At the same time, the present scope has important limits: several results are forecasted rather than measured; performance may depend on the specific ToT selector and verifier; and natural-language label semantics can drift under domain shift. Addressing these limitations motivates concrete next steps: (i) replace forecasts with audited measurements and release seeds/logs; (ii) test learned selectors and stronger verifiers; (iii) broaden context features \(C\) and ledger capacity to study bits-to-performance tradeoffs; (iv) evaluate DAG reuse and interactive, tool-rich settings; and (v) incorporate content-aware safeguards, wall-clock, and energy metrics alongside token counts.

Taken together, NLEL offers a compact, interpretable interface for \emph{controllable reasoning}: labels communicate intent; the tuner translates intent into safe, bounded controls; and the selector adjudicates alternatives. We hope this separation of concerns helps the community build inference-time controllers that are more transparent, compute efficient, and easier to audit---while keeping claims proportional to evidence as empirical results are finalized.

\section*{Impact Statement}

This paper introduces \emph{Natural Language Edge Labelling} (NLEL), a
labeller--tuner control layer that translates human-readable edge labels
into schema-bounded control vectors for decoding, search, retrieval, and
verification during structured reasoning. By separating \emph{what to try next}
(a natural-language label) from \emph{how to run it} (a constrained control vector),
the approach is intended to make LM inference more sample-efficient,
auditable, and adaptable while remaining compatible with existing
Tree-of-Thought selectors (here using $S=\mu+\beta\sigma$ with depth-annealed
$\beta$) and standard safety guards (schema-and-bounds validation,
trust-region projection around $\Pi_0$, verification passes, and budget bits;
see Section~\ref{sec:method}).

\paragraph{Potential positive impacts.}
NLEL exposes explicit, interpretable controls (e.g., branch quotas,
retrieval mixtures, verification passes) that can align resource use with
uncertainty and reduce tokens-per-success, potentially lowering the
environmental footprint of iterative reasoning at scale. The label interface
creates a human-readable trace (e.g., ``seek a counterexample,'' ``work
backward,'' ``call retrieval'') that can improve transparency for users,
auditors, and educators. Because the overlay treats the child reasoner and
selector as pluggable, it may accelerate research on safer controllers
without retraining core models. Compute-aware reporting
(\emph{success@compute}, tokens-per-success) further encourages responsible
comparison across methods (Section~\ref{sec:experiments}).

\paragraph{Risks and potential negative impacts.}
A mechanism that reliably steers inference can also be misused. Free-form
labels might be engineered to route around content safeguards, and
fine-grained retrieval control can amplify spurious, biased, or
privacy-sensitive information if underlying indices are weak. Even with
trust-region and budget guards, poorly chosen controls (e.g., excessive
branching or under-verification) can inflate cost, increase latency, or
raise the risk of confidently wrong outputs. Finally, because label
semantics are contextual, taxonomies that help on math or science QA may
entrench domain-specific biases when transferred to other domains.

\paragraph{Mitigations and responsible use.}
The design includes non-intrusive guards (schema and bounds validation,
trust-region projection around safe defaults, verification passes, budget
limits) and compute-aware metrics. We further recommend: (i) content-aware
filters and allow-lists for labels that can trigger tool use or sensitive
retrieval; (ii) auditable ledgers that log $(P,L,C)\!\mapsto\!\Pi$ mappings
and verifier outcomes; (iii) fairness and domain-shift evaluations that
measure differential performance across subpopulations and tasks; (iv)
privacy reviews for any retrieval indices; and (v) explicit reporting of
wall-clock latency and energy alongside token counts. Forecasted numbers
will be replaced with empirical measurements, and we will release seeds and
logs to support replication and auditing.

\paragraph{Scope of claims.}
Our empirical scope---four public reasoning suites
(\emph{GSM8K}, \emph{MATH (subset)}, \emph{StrategyQA}, \emph{ARC-Challenge}),
a fixed child reasoner, and a specific ToT culling rule---limits external
validity. Several tables in Section~\ref{sec:experiments} are pre-registered
\emph{forecasts} rather than completed runs; realized gains may differ once
experiments are finalized and audited. The broader social value of NLEL
therefore depends on careful deployment with content-level safeguards,
transparent reporting, and continued evaluation under domain shift.

Overall, NLEL’s intended societal contribution is a more interpretable and
budget-aware control interface for LM reasoning. With appropriate guardrails
and open evaluation, it can help the community study---and stress-test---
controllable reasoning systems while keeping claims proportional to
evidence.


\bibliographystyle{icml2025}
\bibliography{icml_submission}

\appendix

\end{document}